\documentclass[letterpaper, conference]{ieeeconf}  

\usepackage{times}
\usepackage{epsfig}
\usepackage{graphicx}
\usepackage{amsmath}
\usepackage{amssymb}
\usepackage{xcolor}
\usepackage{booktabs}
\usepackage{bm}
\usepackage[colorlinks=true,linkcolor=black,anchorcolor=black,citecolor=black,filecolor=black,menucolor=black,runcolor=black,urlcolor=black]{hyperref}

\usepackage{outlines}
\let\labelindent\relax
\usepackage{enumitem}
\setenumerate[1]{label=\arabic*.}
\setenumerate[2]{label=\alph*.}

\usepackage{xspace}

\usepackage[
    style=ieee,
    doi=false,
    isbn=false,
    url=false,
    eprint=false,
    backend=biber,
    natbib=true
    ]{biblatex}

\bibliography{bibliography.bib}

\IEEEoverridecommandlockouts 
\title{\LARGE \bf
Mapping High-level Semantic Regions \\ in Indoor Environments without Object Recognition
}

\author{Roberto Bigazzi$^{1}$, Lorenzo Baraldi$^{1}$, Shreyas Kousik$^{2}$, Rita Cucchiara$^{1}$, and Marco Pavone$^{3}$
\thanks{Roberto Bigazzi was supported by the European Union’s Horizon 2020 research and innovation programme under the MSCA PERSEO No 955778. Lorenzo Baraldi and Rita Cucchiara were supported by the ``Fit for Medical
Robotics'' (``Fit4MedRob'') project, funded by the Italian Ministry of University
and Research.
Shreyas Kousik and Marco Pavone were partially supported by the Toyota Research Institute (``TRI'').
This article solely reflects the opinions and conclusions of its authors and not TRI or any other Toyota entity.}
\thanks{$^{1}$Roberto Bigazzi, Lorenzo Baraldi, and Rita Cucchiara are with the Department of Engineering ``Enzo Ferrari'', University of Modena and Reggio Emilia, Italy {\tt\small \{firstname.lastname\}@unimore.it}}%
\thanks{$^{2}$Shreyas Kousik is with the George W. Woodruff School of Mechanical Engineering, Georgia Institute of Technology, USA {\tt\small shreyas.kousik@me.gatech.edu}}%
\thanks{$^{3}$Marco Pavone is with the Department of Aeronautics \& Astronautics, Stanford University, USA {\tt\small pavone@stanford.edu}}%
}

\def \ie {\emph{i.e.}\xspace}
\def \eg {\emph{e.g.}\xspace}
\def \etal {\emph{et al.}\xspace}
\def \etc {\emph{etc}\xspace}

\providecommand{\R}{\ensuremath \mathbb{R}}

\newcommand{\regtext}[1]{\mathrm{\textnormal{#1}}}

\newcommand{\ts}[1]{\textsuperscript{#1}}
\newcommand{\lbl}[1]{^{\regtext{#1}}}
\newcommand{\str}[1]{{\small \tt{#1}}}

\DeclareMathOperator*{\argmax}{argmax}

\newcommand\sbullet[1][.5]{\mathbin{\vcenter{\hbox{\scalebox{#1}{$\bullet$}}}}}

\newcommand{\rgb}{{\regtext{rgb}}}
\newcommand{\dep}{{\regtext{d}}}
\newcommand{\txt}{{\regtext{txt}}}

\newcommand{\topdown}{{\regtext{top}}}

\newcommand{\obs}{s}
\newcommand{\obsset}{S}
\newcommand{\loc}{x}
\newcommand{\prob}{P}

\newcommand{\regionset}{\mathcal{R}}
\newcommand{\region}{r}

\newcommand{\pose}{p}

\newcommand{\occupancy}{o}
\newcommand{\action}{a}

\newcommand{\loss}{\mathcal{L}}
\newcommand{\feature}{\phi}
\newcommand{\featureset}{\Phi}

\begin{document}

\maketitle

\begin{abstract}
   Robots require a semantic understanding of their surroundings to operate in an efficient and explainable way in human environments.
   In the literature, there has been an extensive focus on object labeling and exhaustive scene graph generation; less effort has been focused on the task of purely identifying and mapping large semantic regions.
   The present work proposes a method for semantic region mapping via embodied navigation in indoor environments, generating a high-level representation of the knowledge of the agent.
   To enable region identification, the method uses a vision-to-language model to provide scene information for mapping.
   By projecting egocentric scene understanding into the global frame, the proposed method generates a semantic map as a distribution over possible region labels at each location.
   This mapping procedure is paired with a trained navigation policy to enable autonomous map generation.
   The proposed method significantly outperforms a variety of baselines, including an object-based system and a pretrained scene classifier, in experiments in a photorealistic simulator.
\end{abstract}

\section{Introduction}\label{sec:intro}
A critical ingredient in robot autonomy is \emph{semantic region mapping}, or the capability to recognize high-level semantics of a robot's surroundings while creating a spatial map. 
In this work, we aim to recognize semantic areas at the region-level of an indoor environment such as a house or an office;~\eg, understanding if the robot is in a bedroom, a living room,~\etc.
This task, which we call Indoor Semantic Region Mapping (ISRM), is often addressed by leveraging object detection to classify a region \cite{hernandez2018object,brucker2018semantic,fernandez2020object,qi2020building,wang2018approach}.
However, such an approach may fail when an object appears in multiple regions (\eg a fridge in a garage), when regions are not easily delineated (\eg an open kitchen attached to a living room), or when regions do not contain many objects (\eg a hallway).
Instead, we explore how to endow robots with a semantic recognition of their surroundings by processing sensor inputs holistically.

We note that semantic region recognition could be considered an instance of the well-defined Computer Vision task of Scene Recognition.
However, directly adapting methods suited to this task is impractical because scene recognition methods are trained on large-scale offline datasets of everyday pictures, which differ significantly from the observations taken from the point of view of a robotic agent~\cite{zhou2017places}.
Furthermore, the category labels in scene-specific datasets are usually very specific and do not contain samples with occlusion or multiple categories.

\begin{figure}[!t]
    \begin{center}
    \resizebox{\linewidth}{!}{
    \includegraphics{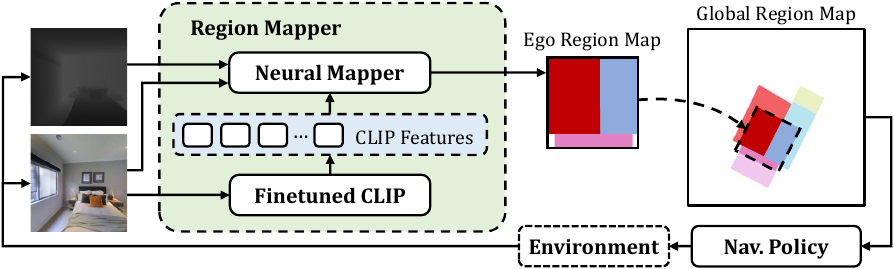}
    }
    \end{center}
    \vspace{-0.3cm}
    \caption{We address the task of indoor semantic region mapping for Embodied Navigation, which requires an agent to build a global semantic understanding of the large-scale regions in an environment.
    Our approach conditions a learned neural mapper with visual features extracted using a region classifier to produce a geometric region map of the environment.
    }
    \vspace*{-0.5cm}
    \label{fig:overview}
\end{figure}

To enable solving ISRM without relying on object recognition, we design a region classification module powered by CLIP~\cite{radford2021learning}, a finetuned, general image-text model.
As CLIP has been trained on a large-scale image-text pair dataset containing natural images, it is generalizable, but it also shows poor performance when dealing with photo-realistic indoor images taken from a simulator.
Therefore, we devise a finetuning strategy with a multi-modal variant of the supervised contrastive loss~\cite{khosla2020supervised}, which allows us to obtain an effective region classifier, even with scarce training data. Further, we design a region mapping architecture that can predict egocentric and global occupancy and region maps, and a navigation module that can exploit the learned region semantics.
An overview of our approach is shown in Fig.~\ref{fig:overview}.
We perform an extensive experimental analysis to compare different approaches to tackle the ISRM task.
Overall, we see this research as a basic step for the next generation of personalized robotics autonomous systems, equipped with vision abilities for moving in complex environments and reasoning at a semantic level to make them suitable for new forms of interaction and cooperation with humans.

\textit{Contributions.}
In summary, we make three contributions.
First, we propose a region classification module that generates grounded language-visual features, suitable for downstream mapping, by using an RGB-D camera and a finetuned CLIP model.
Second, we propose an approach to fusing our region classification features into a global grid map by leveraging an exploration-focused navigation policy~\cite{bigazzi2022focus}.
Third, we provide an extensive set of ablations and baselines to understand the utility and capability of our method, and find that it outperforms an object-based mapping system.
That is, we show that grounding high-level region labels conveys an advantage in robotic semantic mapping.
\section{Related Work}\label{sec:related_work}
We now review relevant work on learning-based mapping, semantic mapping, and scene recognition.

\textit{Learning-Based Mapping.}
A variety of approaches leverage learning to overcome shortcomings in classical SLAM (Simultaneous Localization and Mapping) (\textit{c.f.}, \cite{smith1986representation,thrun2002probabilistic,hartley2003multiple}).
For example, one can learn to generate classical SLAM-style maps~\cite{bigazzi2023embodied}, topological maps~\cite{savinov2018semi,savinov2018episodic}, multi-task deep memory representations~\cite{parisotto2017neural,henriques2018mapnet}, and inferences over unseen regions~\cite{ramakrishnan2020occupancy,landi2022spot,bigazzi2022focus}.
In this work, we propose a learning-based approach to mapping semantic regions.

\textit{Semantic Mapping.}
Semantic mapping approaches can be grouped into two main categories based on generating either low-level or high-level semantic maps.

Low-level semantic maps represent object information and relations, often using scene graphs.
Much of prior work focuses on learning-based methods using preexisting scene graphs~\cite{druon2020visual,zeng2020semantic,du2020learning}.
Works that build a scene graph usually ignore the interactive robotic setting, instead restricting to single observations~\cite{yang2018graph,gu2019scene} or predetermined observation sequences~\cite{armeni20193d, wald2020learning,wu2021scenegraphfusion}.
However, recently, Li~\etal\cite{li2022embodied} built scene graphs with an exploration agent; we similarly build our map online during exploration.
Other approaches to low-level semantic mapping produce metric maps capturing object information~\cite{chaplot2020object,cartillier2020semantic}.
Most similar to our method, a variety of approaches begin by performing low-level mapping (labeling objects), then back out high-level region categories \cite{hernandez2018object,brucker2018semantic,fernandez2020object,qi2020building,wang2018approach,zhang2020indoor,li2022crowdsourcing,seichter2022efficient}.
In this work, we instead generate a region map online without explicitly representing low-level information.

High-level maps depict relations between regions or locations, which is less well-explored in the literature than low-level mapping.
One key recent approach has been to generate hierarchical 3D scene graphs that can also capture room relations \cite{rosinol2021kimera,hughes2022hydra}.
Other approaches focus only on top-down semantic understanding for mobile robot navigation~\cite{narasimhan2020seeing}.
Most similar to our work, Sunderhauf~\etal\cite{sunderhauf2016place} used an external classifier to assign labels to RGB-D observations, then fused these into a semantic map.
In contrast, our approach directly produces the semantic region map end-to-end, and can ground multiple labels spatially within a single observation.
Thus, our work is also similar to the outdoor-focused work of Gan~\etal\cite{gan2020bayesian}, except that we leverage a vision-language model that enables more generalizable semantic mapping, instead of a pretrained semantic segmentation method with a fixed set of classes. 
We also note that such vision-language embeddings show promise for translating language commands into robot navigation trajectories \cite{shafiullah2022clip,shah2022lmnav}, but without creating a semantic map.

\textit{Scene Recognition.}
The task of correctly classifying a scene in an image has captured large interest in the Computer Vision community.
Scene-centric datasets have been released to address this challenge.
For example, Scene15~\cite{oliva2001modeling} was the first testbed for such a task but had a limited number of samples and classes.
MIT Indoor67~\cite{quattoni2009recognizing} and SUN~\cite{xiao2010sun} enlarged the set of classes but the small number of samples limited the application of deep learning.
More recently, Zhou~\etal released Places~\cite{zhou2017places}, containing $\approx$8 million images.
However scene-centric datasets are ill-suited for robot navigation, wherein observations are likely to contain occlusion or be uninformative (\eg facing a wall).
To address these shortcomings, we collect an offline dataset for the classification of rooms inside environments used for embodied navigation.
\section{Problem Statement}\label{sec:problem_statement}

\begin{figure} [t]
    \small
    \centering
    \setlength{\tabcolsep}{.2em}
    \resizebox{.68\linewidth}{!}{
    \begin{tabular}{ccc}
    \textbf{RGB} & \textbf{Depth} & \textbf{Semantic Map} \\
    \addlinespace[0.08cm]
    \includegraphics[width=0.34\columnwidth]{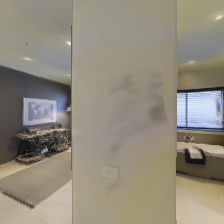} & \includegraphics[width=0.34\columnwidth]{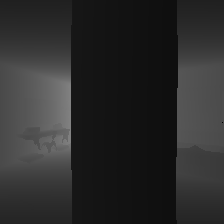} & \includegraphics[width=0.34\columnwidth]{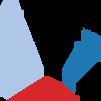} \\
    \addlinespace[0.02cm]
    \includegraphics[width=0.34\columnwidth]{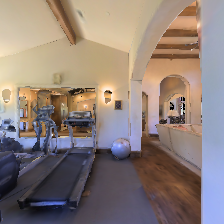} & \includegraphics[width=0.34\columnwidth]{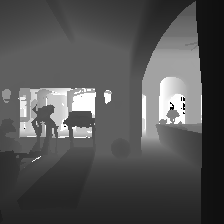} & \includegraphics[width=0.34\columnwidth]{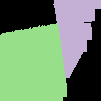} \\
    \addlinespace[0.12cm]
    \end{tabular}
    }
    \resizebox{\linewidth}{!}{
    \begin{tabular}{c}
    \includegraphics{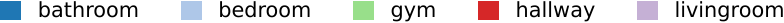} \\
    \end{tabular}
    }
    \vspace{-0.4cm}
    \caption{Sample observations and corresponding top-down semantic maps from the extracted dataset.
    The upper row shows a common challenge faced in the dataset: occlusions and multiple semantic categories in a single image.
    Furthermore, on the right side of the upper image, the region is not obviously recognizable as a bathroom.
    }
    \vspace{-0.5cm}
    \label{fig:sample}
\end{figure}

The goal of this work is to produce top-down semantic maps depicting the different regions of an unseen environment while exploring it online.
This is a challenging task because semantically mapping an unseen environment requires one to model and extract semantic information from the available observations while generalizing over the possible different appearances of each region. 
The goal is to create semantic maps that can be exploited for downstream navigation tasks.
Overall, ISRM consists of two subtasks: \textit{Region Classification} and \textit{Region Mapping}.

\textit{Notation and Setup.}
In the following, we denote a probability distribution over $x$ given $y$ as $\prob(x \mid y)$.
A set of $n$ regions is denoted as $\regionset = \{\region_1,\cdots,\region_n\}$, where the $i$\ts{th} region category is defined as $\region_i \in \mathcal{R}$ (e.g., $\region_1 = \str{bathroom}$).
Similarly, an RGB-D observation at timestep $t$ extracted from the point of view of the agent is denoted as $s_t = \{\obs^\rgb_t, \obs^\dep_t\}$.
The pose (2-D position and heading) of the robot at time $t$ is $\pose_t$, and the action taken at that time is $\action_t$.

\textit{Region Classification.}
Suppose the robot has acquired an RGB observation $\obs^\rgb_t$ of the environment.
In region classification, we seek to model a categorical distribution over regions given an RGB observation, \ie~$\prob(r_i \mid \obs^\rgb_t)$.
Note that this formulation allows for ambiguity in classification, which frequently occurs in the dataset as shown in Figure \ref{fig:sample}.

\textit{Region Mapping.}
The output of region classification is not directly amenable to downstream tasks; hence, we seek to perform region mapping, to create a global map of semantic regions.
In particular, suppose we are given an arbitrary 2D location $\loc$, and a history of observations up to time $t$, denoted $\obsset = (\obs_0,\cdots,\obs_t)$. 
We seek to model a distribution over possible regions, $P(\region_i \mid \loc, \obsset)$, and a distribution over static obstacle occupancy, $P(\occupancy \mid \loc, \obsset)$, at the location $\loc$.
\section{Proposed Method}\label{sec:proposed_method}
In this section, we first introduce the collection procedure and the features of our dataset of observation-map pairs.
Then, we present our method for semantic region mapping and navigation.

\subsection{Dataset}
In order to train a model for region recognition inside photo-realistic environments we extract an off-line dataset of RGB-D observations from the point of view of the agent coupled with the corresponding top-down ground-truth region maps.
Collecting a dataset rather than getting top-down maps directly from the simulator also allows us to efficiently shuffle the training samples and avoid an episode-based online setting, which would increase computational costs and lower the generalization capabilities of the trained method due to forgetting~\cite{mccloskey1989catastrophic,robins1995catastrophic}.

\subsubsection{Extraction Procedure}
The offline dataset is extracted by running a state-of-the-art exploration method~\cite{bigazzi2022focus} using the Habitat simulator~\cite{savva2019habitat} on the environments contained in the Matterport3D dataset for a total of $236$K samples. 
The observation-map pairs are saved taking into consideration the pose of the agent and are stored only if the agent has not visited the same position (within $0.1$ meters) and orientation (within $0.1$ radians $\approx$ $6$ degrees).
We generate exploration episodes using the start position of the episode contained in the object-goal navigation task on Matterport3D dataset \cite{batra2020objectnav}, which are collected on $61$ and $11$ unique building respectively for training and validation splits. 

\subsubsection{Dataset Characteristics} The resulting dataset is divided into 228K and 8K observation-map pairs for training and validation splits, respectively, which is a consequence of the original environment-level division of the Matterport3D dataset. Further, we extract a training-val split consisting of a randomly sampled 5\% of the items in the training set.
The RGB-D observation resolution is $224 \times 224$, with the depth observation that goes from $0$ to $10$ meters. The extracted map is $101 \times 101 \times C$ where $C$ is the number of semantic region categories.
We extract the maps maintaining the semantic labels used in the annotation of Matterport3D dataset~\cite{chang2017matterport3d}.
Two samples from the dataset are shown in Figure \ref{fig:sample}.

\subsection{Region Classification Module}
To tackle the task of correctly classifying the current RGB observation $\obs^\rgb_t$ to a region label $\region_i \in\regionset$, we employ the CLIP image-text model~\cite{radford2021learning}, which has been trained contrastively on large-scale image-text pairs and has shown good classification performance also in zero-shot scenarios. Nevertheless, CLIP still needs to be adapted to the task of region classification and to the distribution of indoor images to work properly, so we also design a finetuning strategy.

Overall, the architecture of the region classification module is composed of an image encoder that processes the RGB observation $\obs^\rgb_t$ to obtain the visual features $\feature^\rgb_t$, and a text encoder that takes as input the set of predefined region labels $\regionset$ and extracts corresponding text features ${{\featureset}}^\txt = [\feature^\txt_0,\cdots,\feature^\txt_R]$.
The predicted region label is derived using the cosine similarity between the observation features $\feature^\rgb_t$ and the text features of each region label ${{\featureset}}^\txt$:
\begin{align}\label{eq:region_classification_by_cosine_similarity}
    \region_i \quad \text{s.t.} \quad i = \argmax_j \frac{\feature\lbl{rgb}_t\sbullet\feature_j\lbl{txt}}{\lVert\feature\lbl{rgb}_t\rVert\sbullet\lVert\feature_j\lbl{txt}\rVert}.
\end{align}
The cosine similarity is used to model $\prob(\region_i \mid \loc, \obsset)$ as a categorical distribution.

\begin{figure} [!t]
    \begin{center}
    \resizebox{\linewidth}{!}{
    \includegraphics{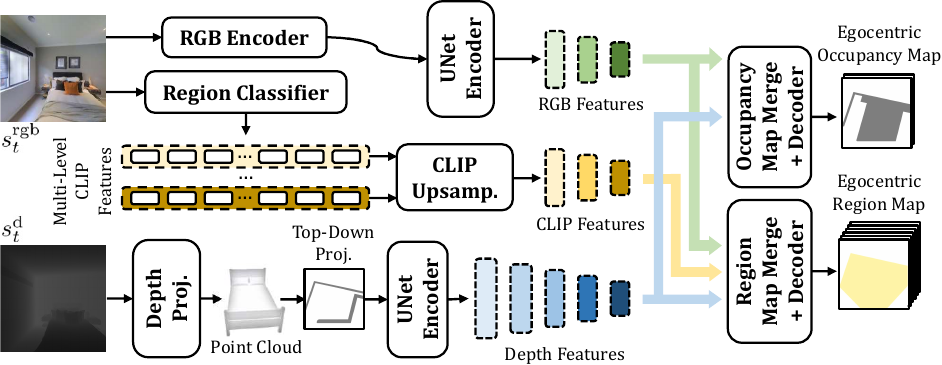}
    }
    \end{center}
    \vspace{-0.3cm}
    \caption{Architecture of the proposed Semantic Region Mapper.
    }
    \vspace{-0.5cm}
    \label{fig:mapper}
\end{figure}

\subsubsection{Multi-Modal Supervised Contrastive Loss}
To lower the computational requirements, we start with pretrained CLIP visual and text encoders and only finetune the projection toward the final embedding space. 
Differently from the image-text matching task on which CLIP has been trained, however, in region classification multiple samples in a mini-batch are likely to belong to the same textual label.
The standard InfoNCE loss is incapable of handling such cases, which can result in reduced performance~\cite{khosla2020supervised}.

To overcome this issue and handle the case of having multiple samples with the same label, we devise a multi-modal adaptation of the Supervised Contrastive Loss~\cite{khosla2020supervised}.
Specifically, given a batch of $N$ image features $\mathcal{V} = [\feature^\rgb_0,\feature^\rgb_1,\cdots,\feature^\rgb_{N-1}]$, we associate to each image the corresponding ground-truth region by using the label associated with the most pixels in the ground-truth top-down semantic map $m^*$. The resulting batch of $N$ labels is used to extract the text features $\mathcal{T} = [\feature^\txt_0,\feature^\txt_1,\cdots,\feature^\txt_{N-1}]$, that are subsequently filtered to remove repeated entries, obtaining $\mathcal{T'} = [(\feature^\txt_0)',(\feature^\txt_1)',\cdots,(\feature^\txt_{K-1})']$. 
Visual and textual features, and their associated labels, are concatenated to obtain the features $\mathcal{X}$ and labels $\mathcal{Y}$ given as
\begin{align}
    \mathcal{X} &= 
        [\feature^\rgb_0,\cdots,\feature^\rgb_{N-1},(\feature^\txt_0)',\cdots,(\feature^\txt_{K-1})'] \nonumber\\
        &= [\feature_0,\cdots,\feature_{N-1},\feature_N,\cdots,\feature_{N+K-1}] \\
    \mathcal{Y} &=
        [y_0,\cdots,y_{N-1},y'_0,\cdots,y'_{K-1}] \nonumber\\
        &= [y_0,\cdots,y_{N-1},y_N,\cdots,y_{N+K-1}].
\end{align}
These are then used in the multi-modal supervised contrastive loss $\loss\lbl{MSC}$:
\begin{align}
\loss\lbl{MSC} = \sum_{i=0}^{N+K-1} -\text{log} \biggl(\tfrac{1}{|B(i)|}\sum_{b \in B(i)} \tfrac{\text{exp}(\feature_i \sbullet \feature_b / \tau)}{\sum\limits_{a \in A(i)}\text{exp}(\feature_i \sbullet \feature_a / \tau)} \biggl),
\end{align}
where $\tau \in \R^+$ is a scalar temperature parameter, $A(i) \equiv \{0, \cdots, N+K-1\} \setminus \{i\}$, $B(i) \equiv \{b \in A(i) : y_b = y_i\}$ is the set of indices of all positives in the batch distinct from $i$, and $|B(i)|$ is its cardinality.

\subsection{Proposed Architecture}
Once we have finetuned our region classification module, we can employ it within a learning-based region mapper paired with a hierarchical navigation policy.
Our method first converts the RGB and depth observations into egocentric occupancy and region maps (\ie distributions over occupancy and regions within the field of view of the robot).
Then, we fuse the egocentric maps to produce a global map.
Fig.~\ref{fig:mapper} shows the architecture of the neural mapper and its modules.

\subsubsection{Egocentric Region Mapping}
The region mapper module builds a region-level semantic map of the environment while generating an obstacle occupancy grid. 
At each timestep, the RGB-D observation $\obs_t = (\obs^\rgb_t, \obs^\dep_t)$ is processed to extract a $L \times L \times (2 + C)$ egocentric map $m_t$ where the first two channels indicate the occupancy and exploration state of the currently observed region, and the last $C$ channels are dedicated to the registration of observed region-level semantic classes.
Each pixel of the map describes the state of a $5 \times 5$ cm area.

The RGB observation $\obs^\rgb_t$ is encoded using a ResNet-18~\cite{he2016deep} encoder followed by a UNet~\cite{ronneberger2015u} encoder.
The depth observation $\obs^\dep_t$, along with intrinsic camera parameters, is used to generate a point cloud of the observed scene where each depth image pixel is projected to its 3D position with respect to the camera.
The point cloud is then collapsed to obtain a top-down view $\obs^\topdown_t$ of the area observed by the agent.
We use another UNet encoder to process the top-down view observation and extract the depth features.

\subsubsection{Semantic Feature Injection}
We employ the region classification model to produce semantically meaningful visual features of the current observation. Specifically, we extract features at different levels of the visual encoder of the region classifier, together with a CNN with transposed convolutional layers to upsample and align the spatial shape of the features to that of the output of the UNet encoders.

At this point, two parallel pipelines follow the generation of the occupancy map and the region map.
We use two different CNNs to merge the multi-modal features obtained from the RGB and depth encoders, and the CLIP features.
The merge module that generates the occupancy map uses RGB and depth features, while the other merge module also includes the CLIP features.
The output of the merge modules is fed to two UNet decoders to produce the final egocentric occupancy map and region map.

Note that the occupancy state that is contained in the first two channels of the map $m_t$ is produced following the method proposed in~\cite{ramakrishnan2020occupancy}.
Therefore, our mapper is not limited to predicting the occupancy map of the visible space but tries to infer also occluded regions of the local map.
However, the prediction of the current region map is limited to the visible area, to avoid projecting incorrect regions behind the walls and the obstacles of the visible region. In order to do so, we use the top-down projection extracted from the depth observation to mask the region map.

\subsubsection{Global Map via Egocentric Map Fusion}
During navigation, the global level map of the environment $M_t$ with dimensionality of $G \times G \times (2 + C)$, where $G > L$, is built using local maps $m_t$ in an incremental fashion.
At each timestep, the pose of the agent $p_t$ is used to apply a corresponding rotation and translation to the local map.
Finally, the transformed local map is fused with the global map $M_t$ using a moving average (we compare against a Bayesian update in Sec.~\ref{sec:experiments}).

\subsubsection{Navigation Module}
Following the literature on embodied exploration in unseen environments, we employ a hierarchical navigation policy that is responsible for predicting the atomic actions of the agent \cite{bigazzi2022focus}.
The policy is composed of a global policy, a deterministic planner, and a local policy.
The global policy takes as input the current state of the global map $M_t$ to generate a long-term goal $g_t$ on the map.
The deterministic planner is in charge of generating the shortest trajectory using the observed occupancy between the agent and the global goal.
A shorter-term goal $l_t$ is sampled on the computed trajectory within $0.25$m from agent's position.
The local policy is trained to reduce the distance from the local goal $l_t$ while avoiding possible obstacles.
Following the hierarchical design, the global goal $g_t$ is updated every $\eta$ timesteps, while the local goal $l_t$ is updated if one of the following conditions holds: the global goal has been updated, the local goal is found to be in an occupied region, or the local goal has been reached.

\subsection{Implementation Details}
\subsubsection{Proposed Architecture}
We tested our model on the environments of Matterport3D using Habitat~\cite{savva2019habitat} simulator with an action space composed of three atomic actions: go ahead $0.25$m, turn left $10^\circ$, and turn right $10^\circ$. The RGB-D observations used the region classification module and by the overall architecture have a $224\times 224$ resolution, the minimum input size for a CLIP model. The mapper module extracts local maps with size $L=101$, while the global maps have size $G=2001$.

\subsubsection{Dataset Region Label Filtering}
Matterport3D contains a large variety of regions, some of which occur infrequently in the data (\eg, \str{bedroom} is common, whereas \str{spa} is not).
Furthermore, there are overlapping region types in the simulator (\eg, \str{porch/terrace/deck}).
To simplify our training procedure and avoid region ambiguities due to overlapping region labels, we postprocess our dataset to only contain the 14 most common labels, including \str{other room}.
We also collapse overlapping region types into a single label.
This results in the following set of filtered labels:
$\regionset = \{$
    \str{bathroom},
    \str{bedroom},
    \str{closet},
    \str{dining room},
    \str{garage}, 
    \str{gym},
    \str{hallway}, 
    \str{kitchen}, 
    \str{library},
    \str{living room},
    \str{office},
    \str{other\ room},
    \str{outdoor},
    \str{stairs}
$\}$.
\begin{table}[t]
\begin{center}
\caption{Accuracy of different CLIP models on room classification on the offline dataset collected in the environments of MP3D.}
\label{tab:offline}
\setlength{\tabcolsep}{.35em}
\resizebox{0.9\linewidth}{!}{
\begin{tabular}{l c cc}
\toprule
\multicolumn{4}{c}{\textbf{Off. MP3D}}\\
\addlinespace[0.8mm]
\textbf{Model} & & {\textbf{Train Val}} & \textbf{Val}\\
\midrule
ResNet50 pretrained & & 29.48 & 26.33 \\
ResNet50 + InfoNCE (only v. proj.) & & 72.22 & 32.18 \\
ResNet50 + MSCL (only v. proj.) & & \textbf{86.85} & \textbf{33.30} \\
ResNet50 + InfoNCE & & 85.66 & 30.34 \\
ResNet50 + MSCL & & 85.87 & 31.02 \\

\addlinespace[0.8mm]

ViT-B/32 pretrained & & 32.07 & 37.53 \\
ViT-B/32 + InfoNCE (only v. proj.) & & 51.16 & 39.63\\
ViT-B/32 + MSCL (only v. proj.) & & \textbf{60.60} & 41.03 \\
ViT-B/32 + InfoNCE & & 52.66 & 43.14 \\
ViT-B/32 + MSCL & & 56.59 & \textbf{44.75}\\

\midrule

Human & & - & $49.58\pm 1.34$ \\

\bottomrule
\end{tabular}
}
\end{center}
\vspace{-0.5cm}
\end{table}

\section{Experiments}\label{sec:experiments}
In this section, we present experiments used to validate the proposed method.
First, we assess the finetuned CLIP model used to condition the region mapper.
Second, we study offline training of the mapper module on our collected dataset.
Finally, we test a state-of-the-art exploration method that is equipped with our pretrained neural region mapper module.

\subsection{Region Classification finetuning}
\subsubsection{Experimental Setup}
We perform an offline finetuning of the room classification module on the collected dataset, and evaluate on two variants of CLIP, namely the ones with a ResNet50 and a ViT-B/32 visual encoder.
As these are two of the most lightweight CLIP variants, they are also appropriate for an embodied setting.
We evaluate in terms of accuracy on the training-val and on the validation split of the collected offline dataset.
The performance on the training-val split, which is a fraction of the training set, is reported to control overfitting. 

\subsubsection{Baselines}
To assess the design of the finetuning methodology, we firstly compare to a baseline using an InfoNCE loss, instead of the proposed multi-modal supervised contrastive loss.
Further, we also experiment when finetuning the final linear projections of both the visual and textual encoder, and when finetuning only the visual projection. In parallel, to give an idea of the difficulty of the task, we tested human performance in a user study dividing the validation unseen of the offline dataset into 16 parts and asking 16 people unaware of the environments in Matterport3D to classify each observation with a region label.

\subsubsection{Results and Discussion}
As shown in Table~\ref{tab:offline}, pretrained CLIP models show a low zero-shot classification performance, with a validation accuracy ranging from 26\% (in the case of ResNet50) to 37.5\% (in the case of the ViT-based backbone).
This highlights that the specificity of the task requires better alignment of the visual appearances associated with each class, and motivates the need for finetuning.
We also note that trying to adapt such a model online with the navigation of the agent could result in poor performance due to the high correlation between observations in an episode.
Therefore, shuffling the training pairs used for finetuning is beneficial for the generalization capabilities of the model. 

Using the multi-modal supervised contrastive loss (MSCL) provides a considerable performance gain, while finetuning the textual encoder only provides better results on one architecture,~\ie ResNet50.
We notice that, while finetuning provides an improvement over the pre-trained backbone, it also suffers overfitting, which might be worth investigating in future architectural and training variants.
Overall, finetuning the visual projection provides a validation accuracy of 33.3\% for ResNet50 and 44.75\% for ViT-B/32, highlighting the benefit of finetuning the image-text model for region classification.

\subsection{Region-Level Mapping Training}
\begin{table}[t]
\begin{center}
\caption{Region-level mapping performance on offline MP3D.}
\label{tab:finetuning}
\setlength{\tabcolsep}{.35em}
\resizebox{\linewidth}{!}{
\begin{tabular}{l c cc c cc}
\toprule
\multicolumn{7}{c}{\textbf{Off. MP3D}}\\
\addlinespace[0.8mm]
& & \multicolumn{2}{c}{\textbf{Train Val}} & & \multicolumn{2}{c}{\textbf{Val}}\\
\cmidrule{2-4}\cmidrule{6-7} 
\textbf{Model} & & \textsf{Acc} ($\uparrow$) & \textsf{IoU} ($\uparrow$) & & \textsf{Acc} ($\uparrow$) & \textsf{IoU} ($\uparrow$)\\
\midrule
ResNet50 pretrained + projection & & 2.22 & 0.19 & & 2.42 & 0.62 \\
ViT-B/32 pretrained + projection & & 3.26 & 0.82 & & 3.96 & 0.62  \\

\addlinespace[0.4mm]

ResNet50 fin. (o.v.p.) + projection & & 14.86 & 7.28 & & 2.81 & 0.62 \\
ViT-B/32 fin. + projection & & 16.81 & 1.64 & & 5.29 & 0.66 \\

\addlinespace[0.4mm]

ResNet50 fin. w. repeated features & & 47.08 & 16.84 & & \textbf{31.41} & 16.45 \\
ResNet50 fin. w. spatial features & & 46.68 & 16.65 & & 31.31 & 15.10 \\

\addlinespace[0.4mm]

ViT-B/32 fin. w. repeated features & & 44.57 & 16.71 & & 31.10 & 17.13 \\
ViT-B/32 fin. w. spat. feat. (Ours) & & \textbf{54.24} & \textbf{28.56} & & 31.19 & \textbf{18.40} \\
\bottomrule
\end{tabular}
}
\end{center}
\vspace{-0.5cm}
\end{table}

\subsubsection{Experimental Setup} 
We then move to the experiments using the region-level mapper evaluating it in terms of pixel-level mapping accuracy and mean Intersection-over-Union with respect to ground-truth region maps extracted from the offline dataset. Also, in this case, the evaluation is conducted on both the train-validation and validation splits.

\subsubsection{Baselines}
We consider three baselines.
First, we consider a simple baseline in which the most probable region class, as predicted by the image-text model, is directly employed to build the egocentric map, without employing the region-map merge component.
In this case, the predicted region class is projected using the depth map directly on the egocentric map.
Furthermore, we report the results obtained when employing the CLIP module finetuned with MSCL as part of our semantic region mapper.
When injecting the features extracted from the CLIP visual encoder, we devise two different alternatives as baselines: our second baseline repeats the final feature vector spatially, to match the spatial shape of the UNet encoders, and our third baseline employs the CNN with transposed convolutions to upsample and match spatial resolutions.

\subsubsection{Results and Discussion}
Results are reported in Table~\ref{tab:finetuning}. First, we observe that repeating the most probable semantic class, as predicted from the egocentric observation of the agent, provides low performance. For instance, a pre-trained CLIP with ViT-B/32 reaches around 4\% accuracy on the validation set. Applying our finetuning strategy increases performance, up to 5\% accuracy on the validation set.
Injecting CLIP features in the semantic region mapper instead provides significantly better results, highlighting that a proper strategy to translate CLIP predictions into an egocentric map is needed.
In this regard, we notice that the spatial features extracted from a ResNet50 are less powerful than those extracted from the ViT-based backbone.
When using the ResNet50 backbone, we notice that simply repeating the global feature vector is better than employing intermediate and spatial-aware features.
On the contrary, the ViT-based backbone provides better spatial features.
Overall, the proposed models achieve up to 31.4\% accuracy and up to 18.4\% IoU on the validation set.

\subsubsection{Qualitative Samples}
\begin{figure}[!t]
    \begin{center}
    \footnotesize
    \setlength{\tabcolsep}{.2em}
    \resizebox{.95\linewidth}{!}{
    \begin{tabular}{cccc}
    \addlinespace[0.08cm]
    \textbf{RGB} & \textbf{Depth} & \textbf{Semantic Map}  & \textbf{GT Sem. Map} \\
    \includegraphics[width=0.25\columnwidth]{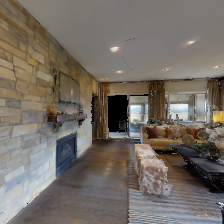} & 
    \includegraphics[width=0.25\columnwidth]{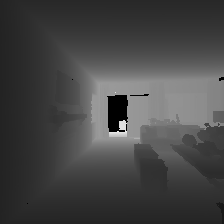} & 
    \includegraphics[width=0.25\columnwidth]{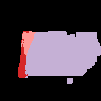} & 
    \includegraphics[width=0.25\columnwidth]{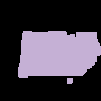} \\
    \addlinespace[0.02cm]
    \includegraphics[width=0.25\columnwidth]{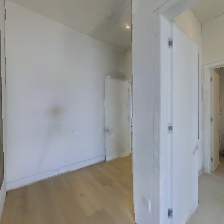} & 
    \includegraphics[width=0.25\columnwidth]{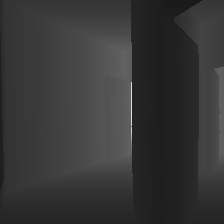} & 
    \includegraphics[width=0.25\columnwidth]{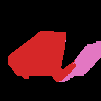} & 
    \includegraphics[width=0.25\columnwidth]{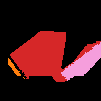} \\
    \addlinespace[0.10cm]
    \multicolumn{4}{c}{\includegraphics[width=.75\columnwidth]{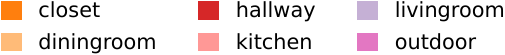}} \\
    \end{tabular}
    }
    \end{center}
    \vspace{-.2cm}
    \caption{Samples of generated maps with corresponding ground truth and visual inputs on the environments of Matterport3D. Note that the first row shows that the model defines the region between the TV and the couches as a \str{hallway} (in red).}
    \label{fig:qualitatives}
\end{figure}

Following the quantitative analysis of the map generation of the region mapper, in Figure~\ref{fig:qualitatives} we present some samples of the output of the region map decoder with the corresponding input acquired on the environments of Matteport3D (MP3D) dataset~\cite{chang2017matterport3d}.

\subsubsection{Ablation Study}
To validate the contribution of using both RGB and depth modalities for the input of the neural mapper, in Table~\ref{tab:off_modalities} we compare our approach using ViT-B/32 backbone and spatial features with two baselines using each modality singularly.
The model using RGB-D surpasses the other baselines, indicating that using both RGB and depth information is beneficial for the final results.

\begin{table}[t]
\begin{center}
\caption{Region-level mapping performance on the offline Matterport3D dataset using different input modalities.}
\label{tab:off_modalities}
\setlength{\tabcolsep}{.35em}
\resizebox{0.77\linewidth}{!}{
\begin{tabular}{l c cc c cc}
\toprule
\multicolumn{7}{c}{\textbf{Off. MP3D}}\\
\addlinespace[0.8mm]
& & \multicolumn{2}{c}{\textbf{Train Val}} & & \multicolumn{2}{c}{\textbf{Val}}\\
\cmidrule{2-4}\cmidrule{6-7} 
\textbf{Model} & & \textsf{Acc} ($\uparrow$) & \textsf{IoU} ($\uparrow$) & & \textsf{Acc} ($\uparrow$) & \textsf{IoU} ($\uparrow$)\\
\midrule
Ours RGB-only & & 27.42 & 15.48 & & 15.75 & 6.52 \\
Ours Depth-only & & 46.34 & 22.83 & & 28.88 & 17.32 \\
Ours & & \textbf{54.24} & \textbf{28.56} & & \textbf{31.19} & \textbf{18.40} \\
\bottomrule
\end{tabular}
}
\end{center}
\vspace*{-0.5cm}
\end{table}

\subsection{Online Mapping Experiments}
\begin{table}[t]
\begin{center}
\caption{Online mapping performance over MP3D dataset.}
\label{tab:online}
\setlength{\tabcolsep}{.35em}
\resizebox{\linewidth}{!}{
\begin{tabular}{l c ccc}
\toprule
\multicolumn{5}{c}{\textbf{MP3D Val}}\\
\addlinespace[0.8mm]
\textbf{Model} & & \textsf{maskAcc} ($\uparrow$) & \textsf{ovrAcc} ($\uparrow$) & \textsf{IoU} ($\uparrow$) \\
\midrule

Baseline w. oracle object detection  (\textit{c.f.} \cite{hernandez2018object,brucker2018semantic,fernandez2020object,qi2020building,wang2018approach}) & & 26.58 & 8.63 & 19.44 \\
Baseline w. pretrained scene detector (\textit{c.f.} \cite{sunderhauf2016place}) & & 28.46 & 9.42 & 21.03 \\

ViT-B/32 (repeated features) & & 33.67 & 18.60 & 20.10 \\

ViT-B/32 (spatial features) and Bayesian update & & 13.08 & 10.58 & 6.48 \\

ViT-B/32 (spatial features) and noise (ours) & & 32.73 & 13.74 & 22.08 \\
ViT-B/32 (spatial features) (ours) & & \textbf{40.81} & \textbf{22.70} & \textbf{24.68} \\
\bottomrule
\end{tabular}
}
\end{center}
\vspace{-0.5cm}
\end{table}

\subsubsection{Experimental Setup}
The online mapping experiment is performed using the Habitat simulator~\cite{savva2019habitat} with Matterport3D (MP3D)~\cite{chang2017matterport3d} environments.
Among the available datasets for photo-realistic embodied navigation~\cite{bigazzi2022embodied} like Gibson~\cite{xia2018gibson}, Habitat-Matterport3D (HM3D)~\cite{ramakrishnan2021hm3d}, only Matterport3D contains room annotations.
We test our mapping approach on board of a state-of-the-art exploration method~\cite{bigazzi2022focus} to evaluate the agent's ability to correctly classify the observed regions.
We compare against a baseline using a perfect object detector that maps the environment by propagating the region labels associated with the closest object characterizing a specific region (\str{bed} $\rightarrow$ \str{bedroom}) similar to \cite{hernandez2018object,brucker2018semantic,fernandez2020object,qi2020building,wang2018approach}.
We also test a baseline based on Sunderhauf \textit{et al.}~\cite{sunderhauf2016place}, but using a ResNet50 model trained on the more recent \str{Places365} dataset~\cite{zhou2017places}.

\subsubsection{Global Room-Level Mapping}
In Table~\ref{tab:online} we evaluate the performance of our mapper in the online setting.
For our approach, we use CLIP with ViT-B/32 backbone, since it outperformed ResNet in the offline setting.
We test both repeated and spatial features.
We also compare a Bayesian map update as opposed to a moving average.
Finally, to test the validity of assuming perfect pose and depth information in our formulation, we also compare with a noisy baseline using Habitat's built-in noise models for the pose sensor~\cite{savva2019habitat} and the depth camera~\cite{choi2015robust}

We report the results in terms of IoU with respect to the entire ground-truth semantic map, the pixel-level accuracy restricted to the area explored by the agent ($\mathsf{maskAcc}$), and the overall pixel-level accuracy over the entire navigable area ($\mathsf{ovrAcc}$).
As expected, the results are generally in line with the offline setting, with the ViT-B/32 backbone with spatial features providing the best results (masked accuracy of 40.81\% and IoU of 24.68\%).
The baseline using oracle object detection achieves only 26.6\% accuracy, since areas close to characteristic objects often belong to different rooms.
We also find that a moving average is more robust than a Bayesian update against spurious and rapidly-changing high-confidence classifications from timestep to timestep (e.g., outputting \str{bedroom} with high confidence at time $t$, then \str{bathroom} at $t+1$), which is a well-known problem with neural network classifiers \cite{nguyen2015deep}.
Finally, we see that noise in pose and depth does produce some impact on performance, but our method with noise still outperforms the other noise-free baselines.
\section{Conclusion}\label{sec:conclusion}
We proposed a method for semantic region mapping for Embodied navigation in indoor environments. 
Our research is motivated by the need to endow agents with a semantic mapping of their surroundings, which is currently under-explored and essential to provide future embodied agents with semantic understanding capabilities.
Our approach comprises a region classification module, based on a finetuned CLIP model, a region mapping architecture with semantic feature injection, and a navigation module, aware of high-level region semantics.
Experiments, conducted on a novel dataset for the region mapping task, and on the Matterport 3D dataset, show that endowing a robot with high-level semantic understanding can convey an advantage over typical low-level object-based mapping.

\printbibliography

\end{document}